\pdfoutput=1

\documentclass[11pt]{article}

\usepackage{acl}

\usepackage{times}
\usepackage{latexsym}

\usepackage[T1]{fontenc}

\usepackage[utf8]{inputenc}

\usepackage{microtype}

\usepackage{multirow}
\usepackage{amsmath}
\usepackage{amsfonts}
\usepackage{amssymb}

\usepackage{booktabs}

\usepackage{paralist}

\usepackage{mdwlist}
\usepackage{graphicx}

\usepackage{color}

\usepackage[ruled,linesnumbered]{algorithm2e}

\usepackage{xspace}

\newcommand{\xx}{\mathbf{x}}

\newcommand{\zz}{\mathbf{z}}
\newcommand{\yy}{\mathbf{y}}

\usepackage{amsfonts,bm}
\newcommand{\loss}{\ensuremath{\mathcal{L}}}
\newcommand{\xvec}{\ensuremath{\mathbf{x}}}
\newcommand{\yvec}{\ensuremath{\mathbf{y}}}
\newcommand{\zvec}{\ensuremath{\mathbf{z}}}
\newcommand{\hvec}{\ensuremath{\mathbf{h}}}

\newcommand{\ie}{\textit{i}.\textit{e}.}

\SetAlCapNameFnt{\small}
\SetAlCapFnt{\small}
\usepackage{threeparttable}

\newcommand{\method}{MoSST\xspace}

\title{Learning When to Translate for Streaming Speech}

\author{Qianqian Dong\textsuperscript{\rm 1}\thanks{\ \ Equal contribution.}, 
 ~Yaoming Zhu\textsuperscript{\rm 1}\footnotemark[1], 
 ~Mingxuan Wang\textsuperscript{\rm 1}, 
 ~Lei Li\textsuperscript{\rm 2} \thanks{~~Work is done while at ByteDance.}\\
\textsuperscript{\rm1} ByteDance AI Lab ~~~~~~~~~~\textsuperscript{\rm2} University of California, Santa Barbara \\
\texttt{\{dongqianqian,zhuyaoming,wangmingxuan.89\}@bytedance.com} \\
\texttt{leili@cs.ucsb.edu}
}

\begin{document}
\maketitle

\begin{abstract}
How to find proper moments to generate partial sentence translation given a streaming speech input?
Existing approaches waiting-and-translating for a fixed duration often break the acoustic units in speech, 
since the boundaries between acoustic units in speech are not even. 
In this paper, we propose \method, a simple yet effective method for translating streaming speech content.
Given a usually long speech sequence, we develop an efficient \textbf{mo}notonic \textbf{s}egmentation module inside an encoder-decoder model to accumulate acoustic information incrementally and detect proper speech unit boundaries for the input in \textbf{s}peech \textbf{t}ranslation task. Experiments on multiple translation directions of the MuST-C dataset show that \method outperforms existing methods and achieves the best trade-off between translation quality (BLEU) and latency. 
Our code is available at \url{https://github.com/dqqcasia/mosst}.

\end{abstract}

\section{Introduction}
\label{sec:intro}

\begin{figure}[!tb]
    \centering
    \includegraphics[width=1\linewidth]{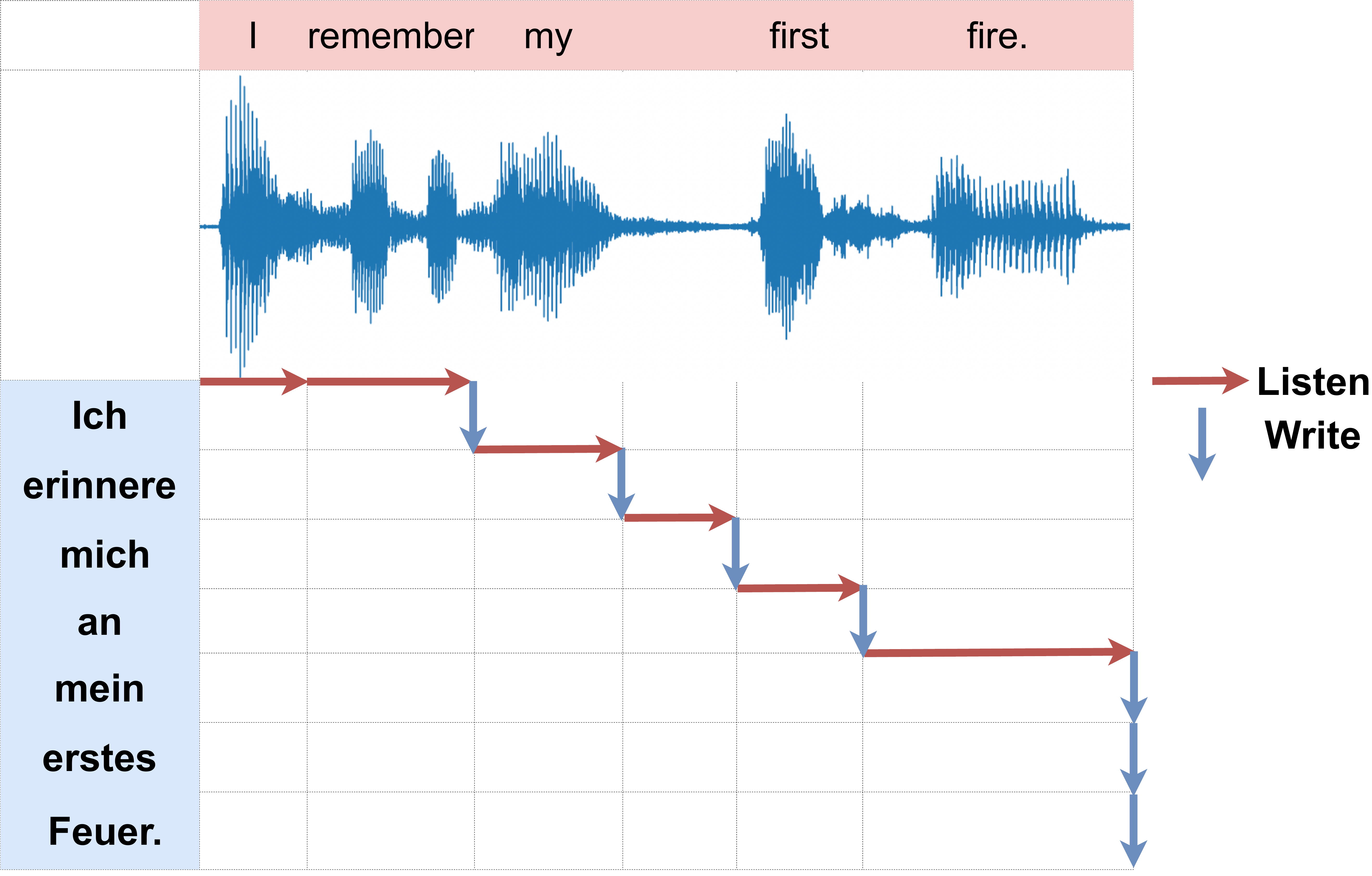}
    \caption{An illustration of streaming speech-to-text translation. ST models listen to the audio in source language, and generate tokens in target language.}
    \label{fig:overview}
\end{figure}

Speech translation (ST) aims at translating from source language speech into target language text, which is widely helpful in various scenarios such as conference speeches, business meetings, cross-border customer service, and overseas travel. 
There are two kinds of application scenarios, including the non-streaming translation and the streaming one. The non-streaming models can listen to the complete utterances at one time and then generate the translation afterward. While, the streaming models need to balance the latency and quality and generate translations based on the partial utterance, as shown in Figure~\ref{fig:overview}.

Recently, end-to-end approaches have achieved remarkable progress in non-streaming ST. Previous work~\cite{weiss2017sequence,berard2018end,bansal2018pre,bansal2019pre,alinejad2020effectively,stoian2020analyzing} ~\citet{ansari2020} has shown that an end-to-end model achieves even better performance compared to the cascaded competitors. 
However, attempts at end-to-end streaming ST are still not fully explored. Traditional streaming ST is usually formed by cascading a streaming speech recognition module with a streaming machine translation module~\cite{oda2014optimizing,dalvi2018incremental}. 
Most of the previous work focuses on simultaneous text translation~\cite{gu2017learning}. \citet{ma2019stacl} propose a novel wait-$k$ strategy based on the prefix-to-prefix framework, which is one of the popular research methods of simultaneous text translation. 
For end-to-end streaming ST,
~\citet{ma2020simulmt,ren2020simulspeech,ma2021streaming} introduce the methodology of streaming machine translation into streaming ST and formalize the task, which belongs to the first study to propose simultaneous ST in an end-to-end manner. 

\begin{figure*}[!t]
    \centering
    \includegraphics[width=0.7\linewidth]{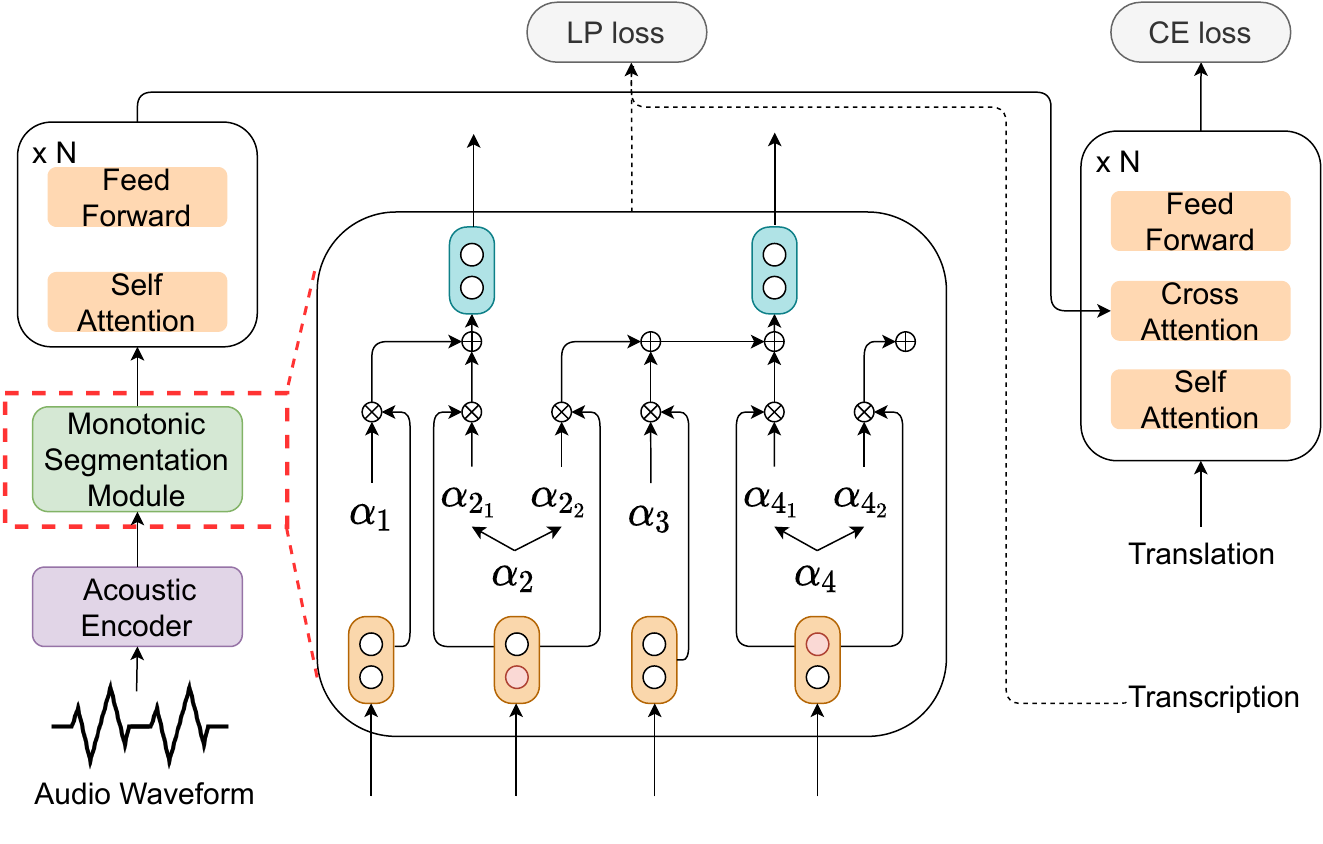}
    \caption{Overview of the proposed \method. \method consists of a pre-trained acoustic model, a monotonic segmentation module (MSM) module, and standard Transformer blocks. The acoustic model extracts features from the raw audio waveform. MSM learns a soft and monotonic alignment over the extracted features from the acoustic model and outputs accumulated acoustic vectors as the input for down-streaming Transformer blocks.}
    \label{fig:model}
\end{figure*}

However, those previous streaming ST systems generally treat a fixed time-span of audio as a acoustic unit and translate new words based on fixed time segmentation, which might be unfavorable for streaming ST translation. Since the speaker's speech speed and the length of the phonemes are distinct, previous methods cannot find the best policy to tell whether to continue reading source audio or translate new words when the source audio is streaming in. 
Hence, we expect the model can determine whether the streaming audio information input is enough to translate new words, similar to the manual simultaneous interpretation. This idea inspires Monotonic-segmented Streaming Speech Translation (\method) system. Specifically, we design a new module that helps to judge the acoustic boundaries of the input audio. We then propose a translation strategy that enables the model to decide whether to read the audio stream or write new tokens given the audio prefix. With the new module and decoding strategy, the model's performance on streaming speech translation has been significantly improved.

We highlight our innovations and findings as follows:
\begin{itemize}
\setlength{\itemsep}{0pt}
\setlength{\parsep}{0pt}
\setlength{\parskip}{0pt}

\item We propose a simple but effective framework, \method for streaming speech translation.
\item We introduce a new monotonic segmentation module to segment audio waveform into acoustic units, based on which we design the adaptive decision strategy which dynamically decides when to translate a new word in streaming scenarios.
\item We validate \method on the MuST-C dataset. The results show that our model significantly outperforms SOTA baselines. Surprisingly, we also find that \method can rival or even surpass other SOTA systems in non-streaming speech translation. Furthermore, we conduct a comprehensive study to analyze the utility of the proposed module and decoding strategy.

\end{itemize}

\section{Proposed Method: \method}
\label{sec:approach}

\label{sec:unistoverview}

This section first formulates the ST task in streaming and non-streaming scenarios. Then, we introduce the detailed architecture of \method, as shown in Figure~\ref{fig:model}. Finally, we give the training and inference strategies of \method for streaming and non-streaming cases.

\subsection{Problem Formulation}

The ST corpus usually contains speech-transcription-translation triples $(\xx, \zz, \yy)$. Specially, $\xx=(x_1,...,x_{T_x})$ is a sequence of acoustic features. $\zz=(z_1,...,z_{T_{z}})$ and  $\yy=(y_1,...,y_{T_{y}})$ represents the corresponding transcription in source language and the translation in target language respectively. Usually, the acoustic feature $\xx$ is much longer than text sequences $\zz$ and $\yy$, as the sampling rate of audio is usually above 16,000 Hz, and each word syllable (about 300 ms) will be recorded by thousands of sampling points. 

The streaming ST model aims to translate instantly when speech audio streams in, that is, given a valid audio prefix $\xx_{< \tau}$, where $\tau$ is the time span of the audio piece, we expect the model can translate enough information $\yy_{< K}$, where $K$ is the maximum number of tokens that the model can translate as time $\tau$,~\ie:
\begin{equation}
\label{eq:formulation}
  \Pr(\yy_{ < K} | \xx_{<\tau}) = \prod_{t=1}^K \Pr(y_t | \xx_{<\tau}, \yy_{<t}; \theta)
\end{equation}
where $\theta$ is the parameters of the streaming ST model. Our goal is to find the best $\theta^*$ that maximizes the $\Pr(\yy_{ < K} | \xx_{<\tau})$ in Eq.~\ref{eq:formulation}.

Note that in our research scenario, we require that the translated piece of the sentence shall not be modified once generated, similar to the settings in simultaneous machine translation~\cite{ma2019stacl}.

\subsection{Model Structure}
\method consists of an acoustic encoder, a monotonic segmentation, and a standard Transformer. 

\noindent \textbf{Acoustic Encoder}
The conventional acoustic encoder using FBANK (log-Mel filterbank, FBANK) as feature extractors faces reduced performance with insufficient training data~\cite{san2021leveraging}, which is especially the case in speech-to-text translation tasks. The FBANK also leads to potential information loss, and may corrupt long-term correlations~\cite{pardede2019generalized}. 

To tackle such problems, we apply the recently-proposed pre-trained acoustic models~\cite{chen2020mam,baevski2020wav2vec} as the feature extractor for \method. Those pre-trained acoustic models learn the speech representation in a self-supervised learning (SSL) way. Since pre-trained acoustic models require only a large amount of unlabeled speech, which also alleviates the corpus shortage of ST tasks. In this paper, we utilize Wav2Vec2~\cite{baevski2020wav2vec} as our 
instance.

\noindent \textbf{Monotonic Segmentation Module}
\label{sec:cif}
The previous speech translation model generally attends the whole audio sequence to the translation tokens with a sequence-to-sequence (seq2seq) framework, which brings two problems: 
\begin{inparaenum}[\it 1)]
\item the model does not learn the alignment between audio and translation explicitly, which may confuse the streaming translation model on whether it has read enough acoustic information when generating the translated text; \item the audio sequences are usually much longer than text sequences, which is computationally demanding for the conventional encoder-decoder speech-to-text model to apply the global attention mechanism. 
\end{inparaenum}
Such high computational cost deviates from the requirements in streaming translation scenarios.

We introduce a Monotonic Segmentation Module~(MSM), to relieve drawbacks of existing models. 
The MSM is inspired by the integrate-and-fire (IF) model~\cite{abbott1999lapicque, dong2020cif,yi2021efficiently}.
Specifically, IF neuron has two modes: \textit{integrate} and \textit{firing}. In integrate mode, the IF neuron dynamically receives signals and accumulates information; when the received information exceeds a certain threshold, IF neuron enters firing mode, at which time it outputs a signal (a.k.a. spiking), where the accumulated state contains information received in the previous integrate phase; and finally, the IF neuron will reset itself and re-enter the integrate mode once the firing mode ends.

In the MSM, we utilize the integrate-and-fire cycle to dynamically locate the boundaries of meaningful speech segments.  
At the integrate mode, the model keeps reading and processing speech frames, while at firing mode the model writes the translated tokens. 
MSM takes the representation from the Acoustic Encoder and uses one of the dimensions as signals for integrate-and-fire.
These signals are passed through a Sigmoid function to produce integration weights.
Once the weights are accumulated to a certain threshold~(e.g. =1.0), the module marks the boundary of the current segment and enters a firing mode.
It then aggregates the rest dimensions of encoder representations according to the weights within this segment. 
These are passed to further processing blocks for WRITE operation. 

The MSM operations are defined as follows: 
\begin{align}
    \alpha_t &= \text{sigmoid}(\bm \hvec_{t,d}) \label{eq:alpha} \\
    \bm l_u &= \sum_t^{\text{S}_u}\alpha'_t \bm \hvec_{t,1:d-1} \\
    \hat{n} &= \sum_t^T \alpha_t \\
    \alpha'_t &= \frac{n^*}{\hat{n}} \alpha_t ,
    \label{sum}
\end{align}
Where $\bm \hvec$ is the acoustic vector as an output of the acoustic encoder, and its subscript denotes the scale value of $\bm \hvec$ at timestamp $t$ and $d$-th dimension (\ie, we use the last dimension as the input of IF neurons).
The Sigmoid value of the scale $\bm \hvec_{t,d}$ is the current weight, denoted as $\alpha_t$. We use the current weight to decide mode conversion from integrate to firing: when the accumulated sum of $\alpha_t$ exceeds the threshold value, the model is believed to have READ sufficient speech signals in this integrate stage, and the IF neural fires the accumulated information $\bm l=(\bm l_1,...,\bm l_u)$ to fulfill one integrate-and-fire cycle. And $\text{S}_u$ represents the firing step corresponding to $\bm l_u$.  

Note that the accumulated information $\bm l$ is calculated as a weighted sum of acoustic vectors $\hvec_t $ at a single integrate stage $t$. We call it as information weight $\alpha'_t$, which helps to scale the amount of information contained in each integrate stage.
We calculate the information weight $\alpha'_t$ by normalizing the current weight $\alpha_t$ with the number of tokens in the corresponding transcription $n^*$, which divides the length of the accumulated acoustic vector $\hat{n}$. 

\noindent \textbf{Transformer block}
The last module of the \method is the standard Transformer. The Transformer blocks take the integrated acoustic vector $\bm l$ from the MSM layer as the input, which aims to extract the semantic feature ($\hvec^{SE}$) of the input audio. Since MSM has significantly compressed the length of acoustic features, the Transformer can attend the input and output directly without the excessive computational overhead. Note that to ensure that MSM learns the correct length of acoustic units, we use the length of the corresponding transcription as a supervised signal and introduce length penalty loss (``LP loss'' in Figure~\ref{fig:model}) to assist MSM's learning.
\begin{equation}
    \loss_{\rm lp}(\theta;\xvec,\zvec) = ||n^* - \hat{n}||_2,
\end{equation}
During inference, an extra rounding operation is applied on $\hat{n}$ to simulate $n^*$. Based on the matched sequence length, the accumulated acoustic vector $\bm l$ is mapped back into the model size by a randomly initialized fully connected layer.

\subsection{Training Strategies}

\noindent \textbf{Multi-task Joint Training with ASR}
\method jointly fulfills the ST and ASR tasks with the multi-task learning (MTL) strategy as its main model.
To distinguish two tasks, we add two special task indicators at the beginning of the text as the \emph{BOS} operator for decoding.
For example, if the audio input for "\texttt{Thank you .}" is in English, for ASR, 
we use \texttt{[en]} as the \emph{BOS} and decode $\zz$= "\texttt{[en] Thank you .}". We add \texttt{[De]} at the start of German translation, thus $\yy$ is "\texttt{[De] Danke .}"

Both ST and ASR are optimized with cross-entropy (``CE loss'' in Figure~\ref{fig:model}) losses, defined in Equation~(\ref{eq:st_loss}) and (\ref{eq:asr_loss}) respectively.

\begin{equation}
\loss_{st}(\theta;\xvec,\yvec)=-\sum_{i=1}^{T_y}\log p_{\theta}(y_i|y_{\text{\textless} i}, \hvec^{SE})
\label{eq:st_loss}
\end{equation}
\vspace{-2ex}
\begin{equation}
\loss_{asr}(\theta;\xvec,\zvec)=-\sum_{i=1}^{T_z}\log p_{\theta}(z_i|z_{\text{\textless} i}, \hvec^{SE})
\label{eq:asr_loss}
\end{equation}
where the decoder probability $p_{\theta}$ is calculated from the final softmax layer based on the output of the decoder. 

We use the joint training strategy to optimize all modules.
The overall objective function is the weighted sum for all aforementioned losses:

\begin{equation}
\begin{aligned}
\loss(\theta;\xvec,\yvec,\zvec) = & \alpha\loss_{\rm lp}(\theta;\xvec,\zvec)+\beta\loss_{ce}
\label{eq:overall_loss}
\end{aligned}
\end{equation}

Where $\loss_{ce}$ represents $\loss_{asr}$ or $\loss_{st}$. In the following experimental sections, $\alpha$ is set to 0.05, and $\beta$ is set to 1 by default.

\subsection{Inference Strategies}
\label{sec:infer_strategy}

\noindent \textbf{Wait-$k$ Policy} 
\method adopts wait-$k$ policy for streaming translation, which originates from simultaneous machine translation~\cite{ma2019stacl}. Wait-$k$ policy waits for K source tokens and then translates target tokens concurrently with the source streams in (\ie, output $N$ tokens when given $N+K$ source tokens). 

The previous online ST systems adopt \textit{Pre-fix Decision}~\cite{ma2020streaming,ma2020simulmt} for wait-$k$ policy, where a fixed time span (usually 280ms) of the source waveform is regarded as a new unit. However, the pre-fixed decision is limited on real-world scenarios since the speech speed of speakers and the length of acoustic units are distinct, where a fixed time stride guarantees neither sufficient information if the phonemes are too long, nor a proper translation latency if the phonemes are too short.

\begin{algorithm}[!t]
\small
 \caption{\small{\textbf{Adaptive Decision Strategy}}}
 \label{Alg:policy}
\SetAlgoLined
\KwIn{The waveform sequence $\xx$, the MSM model $\mathbf{M}$, wait lagging $K$}
\KwOut{The translated sentence $\hat{\yy}$}
initialization: the read waveform segment $\hat{\xx} = []$, the output sentence $\hat{\yy} =[]$\;
 \While{$\hat{\yy}_{i-1}$ is not EndOfSentence }{
  calculate MSM integrated state $ \bm l_u$ \;
  \uIf{
  $\hat{\xx} ==  \xx$ \; 
  }
  {
  \tcc{the waveform is finished}
  \tcc{write new token}
  $\hat{\yy}=\hat{\yy}+\text{decoder.predict()}$\;
  M.decoder.update($\hat{\yy}$) \;
   }
   \uElseIf{
   $ |\bm l_u | - | \hat{\yy} | < K $  \;
  }{
  \tcc{read waveform} 
  \label{op:READ}
  $\hat{\xx} = \hat{\xx}+{ \text{new\_segment}(\xx)}$\;
  M.encoder.update($\hat{\xx}$)
  }
  \uElse{
  \tcc{write new token} 
  \label{op:WRITE}
 $\hat{\yy}=\hat{\yy}+\text{decoder.predict()}$\;
  M.decoder.update($\hat{\yy}$) \;
  }
 }
 return $\hat{\yy}$ \;

\end{algorithm}

\noindent \textbf{Adaptive Decision} 
We propose a new decision strategy for streaming speech translation, namely \textit{Adaptive Decision}. Our new strategy dynamically decides when to write the new token according to the integrated state length of MSM (\ie, $ \bm |\bm l_u| $ in Equation (~\ref{sum}) ).

Since MSM scales up the acoustic information monotonically, the model can estimate the acoustic boundary for each units in the audio. We use such integrate feature as a basis to tell whether the information carried by the waveform segment is sufficient; hence the proposed adaptive decision revises the drawbacks in fixed decision. 

We propose our new decoding policy in Algorithm~\ref{Alg:policy}. The new policy utilizes wait-$k$ to decide when to write new translation tokens and adaptive decisions to decide how long the input is regarded as a unit.  
Specifically, during the online ST translation, the model shall decide whether to read new audio frames or translate a new word at any time, called the READ/WRITE decision. We denote $\hat{\xx}$ as the audio sub-sequence that the model has READ from the source and $\hat{\yy}$ as the sentence prefix that has been already generated.
The wait-$k$ policy makes the READ/WRITE decision according to the length difference between the MSM integrated state $\bm |\bm l_u| $ and the generated sentence $| \hat{\yy} |$. 
When the integrated state $\bm |\bm l_u| $ is $K$ word behind the generated $| \hat{\yy} |$, the \method generates a new token (line~\ref{op:WRITE}) and updates decoder states recursively, otherwise, the model waits and reads the audio streaming (line~\ref{op:READ}), and updates the encoder states.

\noindent \textbf{Train-full Test-k}
Streaming translation needs to predict the output based on part of the input. 
If the \textit{train-full test-k} paradigm is applied, the streaming performance will decrease a lot due to the mismatch between training and inferring.
The previous streaming work generally uses a prefix-to-prefix training framework~\cite{ma2019stacl}, implemented by a unidirectional encoder and decoder, and equipped with the waik-$k$ policy. 
In \method, the learned monotonic segmentation module allows our model to have streaming decoding capability without performance drop. 

\section{Experiments}
\label{sec:exps}

\subsection{Dataset}

\textbf{MuST-C}\footnote{\url{https://ict.fbk.eu/must-c/}}~\citep{di2019must} is a multilingual ST corpus with triplet data sources: source audio, transcripts, and text translations. 
To the best of our knowledge, MuST-C is currently the largest ST dataset available. It includes data from English TED talks with auto-aligned transcripts and translations at the sentence level. We mainly conduct experiments on English-German and English-French language pairs.  And we use the dev and tst-COMMON sets as our development and test data, respectively.
\begin{table}[!h]
    \centering
    \small
    \setlength\tabcolsep{1.8pt}
    \begin{tabular}{ccccccc}
    \toprule
     $k$ & 1 & 3 & 5 & 7 & 9 & inf\\
     \midrule
      SimulSpeech~$^\dagger$   &10.73&15.52&16.90&17.46&17.87&18.29 \\
       \textbf{\method}  &\textbf{11.76} &\textbf{15.57}&\textbf{18.23}&\textbf{18.71}&\textbf{19.37}&\textbf{19.95} \\
      \bottomrule
    \end{tabular}
    \caption{Comparison with SimulSpeech on MuST-C EN-DE tst-COMMON test set. $^\dagger$ represents results from ~\citet{ren2020simulspeech}.}
    \label{tab:vs_simulspeech}
\end{table}
\subsection{Preprocessing} 

For speech input, the 16-bit raw wave sequences are normalized by a factor of $2^{15}$ to the range of $[-1, 1)$. For text input, on each translation pair, all texts (including transcript and translation) are pre-processed in the same way. Texts are case-sensitive. We keep and normalize the punctuations, but remove non-print characters. We tokenize sentences with Moses tokenizer \footnote{\url{https://github.com/moses-smt/mosesdecoder/blob/master/scripts/tokenizer/tokenizer.perl}} and filter out samples longer than 250 words.
For subword modeling, we use a unigram sentencepiece~\cite{kudo2018sentencepiece}  with a dictionary size of 10000. On each translation direction, the sentencepiece model is learned on all text data from ST corpora.

\subsection{Model and Experimental Configuration}
\noindent \textbf{Model Configuration}~
For audio input, the Wav2Vec2 module follows the base\footnote{\url{https://dl.fbaipublicfiles.com/fairseq/wav2vec/wav2vec_small.pt}} configuration in ~\citet{baevski2020wav2vec}. It uses parameters self-supervised pre-trained on LibriSpeech audio data only.  
The subsequently shared Transformer module has a hidden dimension of 768 and 4 attention heads. The encoder is 8 layers, and the decoder is 6 layers.
We use the simplified version of the Continuous IF implementation ~\cite{yi2021efficiently} for MSM module, which introduces no additional parameters except for a fully connected layer.

\begin{figure*}[!t]
    \centering
    \includegraphics[width=1\textwidth]{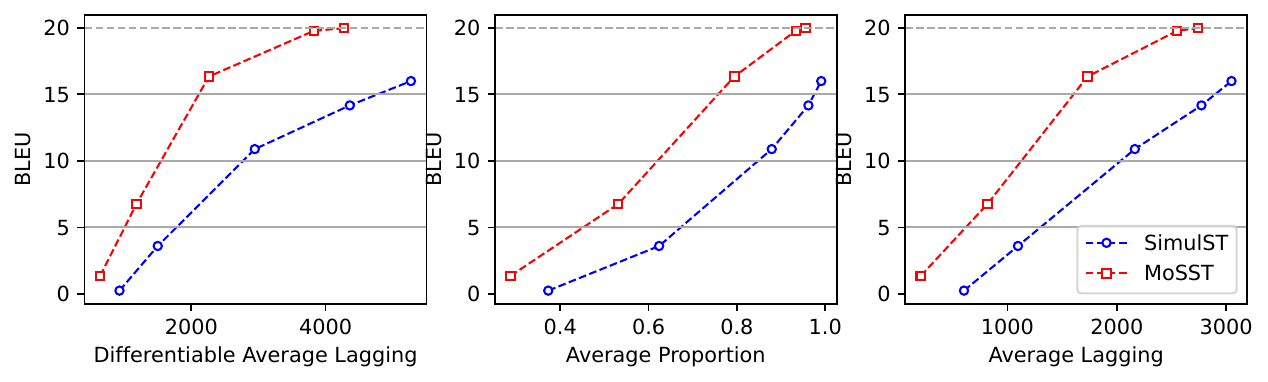}
    \caption{The translation quality against the latency metrics (DAL, AP and AL) on the tst-COMMON set of MuST-C En-De dataset. Decoding strategy here is pre-fixed decision. $k$ in SimulEval is set to 5 as default. The result of SimulST is reproduced by Fairseq~\footnote{\url{https://github.com/pytorch/fairseq/blob/main/examples/speech_to_text/docs/simulst_mustc_example.md}}.
    }
    \label{fig:bleu_latency}
\end{figure*}
\footnotetext[4]{\url{https://github.com/pytorch/fairseq/blob/main/examples/speech_to_text/docs/simulst_mustc_example.md}}

\noindent \textbf{Experimental Configuration}~
We use an Adam optimizer with $\beta_1=0.9, \beta_2=0.98$, and 4k warm-up updates. We set the maximum training batch of the waveform audio token to be 3.2 million.
We apply an inverse square root schedule algorithm for the learning rate. We average 10 consecutive checkpoints around the one with the best dev loss and adopt a beam size of 5. We implement our models in Fairseq~\cite{ott2019fairseq}.

\subsection{Evaluation}
For offline translation, the model's performance is mainly evaluated with quality metrics. While for streaming translation, ST model is evaluated by the latency-quality trade-off curves.  

\noindent \textbf{Quality Metrics}~We quantify translation accuracy with detokenized BLEU~\citep{papineni2002bleu} using sacreBLEU \footnote{\url{https://github.com/mjpost/sacrebleu}}.

\noindent \textbf{Latency Metrics}~Existing simultaneous translation work mainly focuses on the latency evaluation of text translation, and has proposed computation unaware metrics, such as Average Proportion (AP)~\cite{cho2016can}, Average Latency (AL)~\cite{ma2019stacl}, Continues Wait Length (CW)~\cite{gu2016learning} and Differentiable Average Lagging (DAL)~\cite{cherry2019thinking}. \citet{ma2020simuleval} extends the latency metrics of text translation into ST, including AL, AP, and DAL. The latency metrics for streaming \method are evaluated by AL, DAL, and AP based on the SimulEval toolkit~\footnote{\url{https://github.com/facebookresearch/SimulEval}}~\cite{ma2020simuleval}. 

\subsection{Experimental Results}

\subsubsection{Streaming Speech-to-text Translation}

We compare the performance of our method with published work on streaming ST tasks. 
SimulST~\cite{ma2020simulmt} introduces the wait-$k$ training strategy in simultaneous text translation into simultaneous ST tasks. The comparison result on MuST-C EN-DE tst-COMMON set is shown in Figure~\ref{fig:bleu_latency}. It can be seen that \method is significantly better than the baseline system in all the three latency metrics and the quality metric. 
SimulSpeech~\cite{ren2020simulspeech} also adopts the wait-$k$ strategy and leverages the connectionist temporal classification (CTC) decoding to split the input streaming speech chunk in real-time. Besides, SimulSpeech introduces attention-level and data-level knowledge distillation (KD) to improve performance. The comparison result on MuST-C EN-DE tst-COMMON set is shown in Table~\ref{tab:vs_simulspeech}. 
It can be seen that when $k$ ranges from $1$ to infinite, our method significantly outperforms SimulSpeech. 
Existing work all uses the wait-$k$ training strategy implemented with a unidirectional mask, which would damage the performance of offline evaluation in full context. While \method can serve well both non-streaming and streaming translation. At the same time, the shrinking mechanism based on the MSM can speed up model convergence, which we give a detailed analysis in Sec~\ref{sec:training_time} of the Appendix.

\subsubsection{No-streaming Speech-to-text Translation}

We also compare the performance of our method with published work on offline ST tasks under experimental settings without external supervised training data. The result is shown in Table~\ref{table:offline}. Fairseq~\cite{wang2020fairseq}, ESPnet~\cite{inaguma2020espnet}, and NeurST~\cite{zhao2020neurst} are recently emerging R\&D toolkits for ST. Transformer ST uses a standard SpeechTransformer~\cite{dong2018speech} model structure, with a pre-trained ASR model to initialize the encoder and a pre-trained MT model to initialize the decoder. \citet{zhang2020adaptive} propose adaptive feature selection (AFS) for ST, which applies $\mathcal{L}_0$\textsc{Drop}~\cite{zhang2020sparsifying} to dynamically estimate the importance of each encoded speech feature. STAST~\cite{liu2020bridging} uses a speech-to-text adaptation method to bridge the modality gap in the semantic space by MTL and representation regulation with MT.~\citet{le2020dual} adapt the dual-decoder transformer with a dual-attention mechanism to joint ASR and ST for both bilingual (BL) and multilingual (ML) settings. Compared with the best results published so far, \method can achieve an improvement of 1.3 BLEU and 0.7 BLEU respectively. It should be noted that the previous methods can be integrated into \method to expect better performance. We will leave it for further exploration.

\begin{table}[!t]
\centering
\small

\setlength{\tabcolsep}{1mm}{
\scalebox{0.9}{
\begin{tabular}{lcc}
\toprule
\multirow{2}{*}{\textbf{Model}} & \multicolumn{2}{c}{\textbf{MuST-C EN-X}}  \\

& EN-DE & EN-FR\\
\midrule

Transformer ST Fairseq ~\cite{wang2020fairseq}& 22.7 & 32.9 \\
Transformer ST ESPnet ~\cite{inaguma2020espnet}& 22.9 & 32.8 \\
Transformer ST NeurST ~\cite{zhao2020neurst}& 22.8 & 33.3 \\
AFS ST ~\cite{zhang2020adaptive}& 22.4 & 31.6 \\
STAST~\cite{liu2020bridging}~ & 23.1 & -\\
Dual-Decoder Transformer (BL)~\cite{le2020dual}& 23.6 & 33.5\\

Wav2Vec2 + Transformer ~\cite{han2021learning}& 22.3 & 34.3 \\
W-Transf~\cite{ye2021end} & 23.6& 34.6\\
RealTranS~\cite{zeng2021realtrans} & 22.99 & - \\

\textbf{\method} &\textbf{24.9}$^\dagger$  & \textbf{35.3}$^\dagger$ \\
\bottomrule
\end{tabular}
}
}
\caption{Results of non-streaming ST models on MuST-C EN-DE and EN-FR tst-COMMON test set. 
$^\dagger$ indicates our improvement is statistically significant.
}
\label{table:offline}

\end{table}

\section{Analysis}

\subsection{Ablation Studies}
\begin{table}[!t]
\renewcommand\arraystretch{1.1}
\centering
\small
\begin{tabular}{lccccc}
\toprule
\multirow{2}{*}{\textbf{Model}} & \multicolumn{2}{c}{\textbf{EN-DE}}  & & \multicolumn{2}{c}{\textbf{EN-FR}}\\
\cline{2-3}
\cline{5-6}
& BLEU & $\bigtriangledown$ && BLEU & $\bigtriangledown$\\
\midrule
    \textbf{\method} &\textbf{24.9} &- &&\textbf{35.3}&-\\
    ~~w/o MSM &22.7 &-2.2 &&34.4 &-0.9\\
    ~~w/o MTL &21.9 &-0.8  &&33.8 &-0.6\\ 
    ~~w/o Pretrain & 20.0 &-1.9&&31.6&-2.2\\
    \bottomrule
    \end{tabular}
    \caption{Results of ablation study on MuST-C EN-DE and EN-FR tst-COMMON test set. ``w/o MSM'' stands for \method augmented without the monotonic segmentation module in the acoustic encoder.``w/o MTL'' means removing multi-task joint learning with ASR task. ``w/o Pretrain'' represents using FBANK as input instead of the self-supervised acoustic representation.}
    \label{tab:ablation}
\end{table}

We conduct ablation studies to demonstrate the effectiveness of the design of \method, including the monotonic segmentation module, multi-task joint training with ASR, and self-supervised acoustic representation. The ablation study results can be seen in Table~\ref{tab:ablation}. The translation quality decreases significantly when each of the modules or strategies is emitted successively. The self-supervised acoustic representation can bring almost 2 BLEU on both EN-DE and EN-FR datasets, which shows that large-scale SSL brings hope to solving the data scarcity problem of end-to-end ST. For EN-DE language pair, joint training with the auxiliary ASR task has a performance gain of 0.8 BLEU. And the monotonic segmentation module has an additional 2.2 performance gain to our method. The results show a consistent performance improvement on EN-FR language pair. This verifies the outstanding advantage of the monotonic soft attention mechanism of 
MSM in extracting contextual acoustic representations. 

\subsection{Effects of Decoding Strategy}
\begin{figure}[!t]
    \centering
    \includegraphics[width=0.5\textwidth]{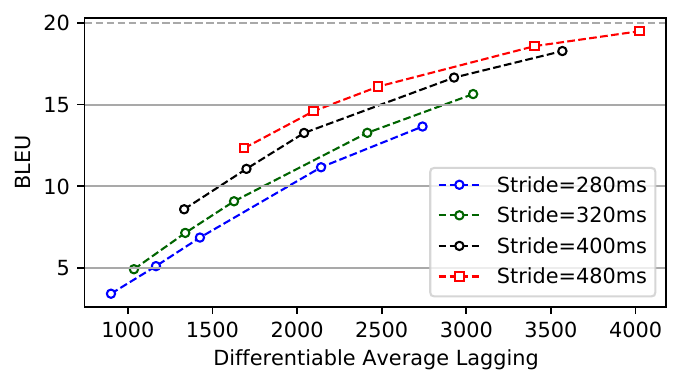}
    \caption{The translation quality against the latency metrics (DAL) on the tst-COMMON set of MuST-C En-De dataset.  Decoding strategy here is pre-fixed decision. Points on the curve correspond to $k$ in SimulEval with 5, 7, 9, 15 and 20, respectively.}
    \label{fig:prefix_decision}
\end{figure}

\subsubsection{Pre-fix Decision}
For the pre-fixed decision decoding strategy, the parameter setting of stride is very important. In Figure~\ref{fig:prefix_decision}, we compare the influence of different strides on the pre-fixed decision strategy. It can be seen that increasing stride within a certain range will have a positive impact on the latency-bleu trade-off. But the model also tends to fall into the field of a larger latency.

\subsubsection{Adaptive Decision}
\begin{figure}[!t]
    \centering
    \includegraphics[width=0.5\textwidth]{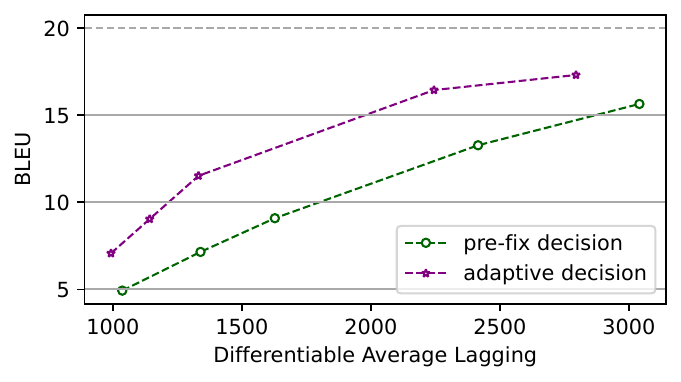}
    \caption{The translation quality against the latency metrics (DAL) on the tst-COMMON set of MuST-C En-De dataset. Pre-fixed decision is tested with the stride size of 320ms.}
    \label{fig:adaptive_decision}
\end{figure}

We have proposed an adaptive decision in Section~\ref{sec:infer_strategy}. 
To better emphasize the latency factor, we compare the performance of the adaptive decision and the pre-fixed decision on the tst-COMMON test subset of MuST-C EN-DE.
The results are shown in Figure~\ref{fig:adaptive_decision}. 
Compared with the pre-fixed strategy decoding method, the adaptive strategy decoding method has a better balance between delay and quality. Through observation, it is found that the adaptive strategy can 
ignore the silent frames. For example, after predicting a punctuation, it will read continuously to accumulate enough source acoustic information. In addition, the adaptive strategy can further reduce the delay by setting the number of WRITE operations after the accumulated information is sufficient according to the length ratio of the source sentences and the target sentences between different language pairs, which requires further exploration.

\subsubsection{Alignment Visualization}

\begin{figure*}[!h]
    \centering
    \includegraphics[width=1\textwidth]{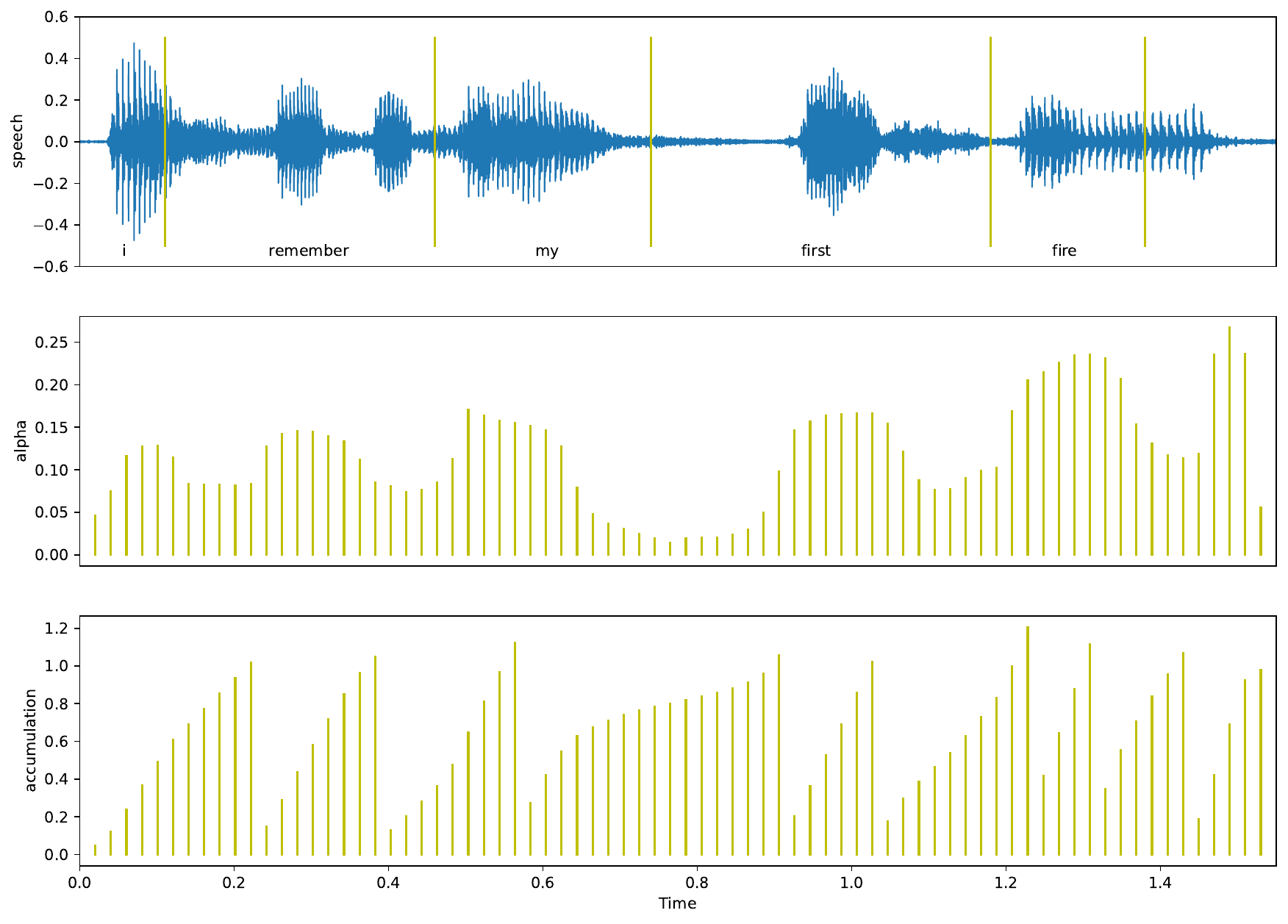}
    \caption{An example speech and its corresponding learned firing positions by \method. 
    Top: the waveform of speech sequence and the true segmentation boundary corresponding to each word.  
    Middle: the learned weights (corresponding to $\alpha$ in Eq.\eqref{eq:alpha}) using the last dimension from acoustic encoder states.
    Bottom: the integrated weights by MSM in \method. MSM will fire and start to compute the segment's representation once the accumulated weights exceed a certain threshold (=1.0).}
    \label{fig:cif_visualization}
\end{figure*}

In Figure~\ref{fig:cif_visualization}, we show the ground truth alignment and the predicted firing positions learned by \method. We can see that what MSM learned is the acoustic boundary, not to mimic wait-$k$. Therefore, the length of the audio chunk can be adaptively read in during streaming decoding, while ensuring that each chunk includes a complete acoustic unit. 

\section{Related Work}
\label{sec:related}

\noindent\textbf{Speech Translation}~
\citet{berard2016listen} have given the first proof of the potential for end-to-end speech-to-text translation without using the intermediate transcription. 
The training method based on pre-training~\cite{weiss2017sequence,berard2018end,bansal2018pre,bansal2019pre,alinejad2020effectively,stoian2020analyzing,dong2021consecutive} can effectively use pre-trained models with better performance as initialization to speed up the convergence of the ST model. Multi-task learning~\cite{weiss2017sequence,berard2018end,liu2020synchronous,indurthi2020end,han2021learning,ye2021end} can fully optimize the model parameters and improve the performance with the aid of auxiliary tasks. Knowledge distillation has been proved to be efficient to learn from pre-trained models~\cite{liu2019end,liu2020bridging,dong2021listen}. 
\citet{DBLP:conf/acl/LePWGSB20} introduce adapter for multilingual speech translation.
Similarly, ~\citet{kano2017structured,wang2020curriculum} introduce curriculum learning methods, including different learning courses of increasing difficulty. To overcome data scarcity, ~\citet{jia2019leveraging,pino2019harnessing} augment data with pseudo-label generation, and ~\citet{bahar2019using,di2019data,mccarthy2020skinaugment} introduce noise-based spectrum feature enhancement. ~\citet{zhang2020adaptive} propose adaptive feature selection to eliminate uninformative features and improve performance. 

\noindent\textbf{Streaming Speech Translation}~
Traditional streaming ST is usually formed by cascading a streaming ASR module and a streaming machine translation module~\cite{oda2014optimizing,dalvi2018incremental}. The ASR system continuously segments and recognizes the transcription of the audio segment, and then the machine translation system continuously translates the text segment output from the upstream. 
Most of the previous work focuses on simultaneous text translation~\cite{gu2017learning}. \citet{gu2017learning}~ learn an agent to decide when to read or write. 
\citet{ma2019stacl} propose a novel wait-$k$ strategy based on the prefix-to-prefix framework to synchronize output after reading $k$ history tokens. Many following work propose some improvement strategies based on adaptive wait-$k$~\cite{zheng2019simpler,zhang2020learning,zhang2020dynamic} and efficient decoding~\cite{elbayad2020efficient,zheng2020opportunistic}. Some monotonic attention methods~\cite{arivazhagan2019monotonic,ma2019monotonic,schneider2020towards} have been proposed to model the monotonic alignment of input and output.
~\citet{arivazhagan2020ret,arivazhagan2020re} propose a re-translation strategy, allowing the model to modify the decoding history to improve the performance of streaming translation.
~\citet{ma2020simulmt} propose SimulST, which applies the wait-$k$ method from streaming machine translation~\cite{ma2019stacl} into streaming ST. ~\citet{ren2020simulspeech} propose SimulSpeech, which uses knowledge distillation to guide the training of the streaming model and the connectionist temporal classification (CTC) decoding to segment the audio stream in real-time. 
\citet{ma2021streaming} enable the streaming model to handle long input by equipping with an augmented memory encoder. 
\citet{chen2021direct} use a separate and synchronized ASR decoder to guide the ST decoding policy. 
\citet{DBLP:conf/acl/ZengLL21} introduce a blank penalty to enhance performance in simultaneous scenarios.

\section{Conclusion}
\label{sec:conclusion}
We propose \method, a simple and effective framework for online speech-to-text translation. 
\method consists of a pretrained acoustic model, a monotonic segmentation module, and a standard Transformer, along with the multitask training strategy and the adaptive decision strategy. 
The monotonic segmentation module and the adaptive decision strategy tell our method when to translate. 
Moreover, the pre-trained acoustic encoder and the multitask training strategy boost our method's ability to predict what to generate. 

The experiment on MUST-C datasets validates the effectiveness of \method over previous work. The results show that \method can achieve a better trade-off between quality and latency over prior end-to-end models and cascaded models in diverse latency settings. Besides, we also find \method can rival non-streaming speech translation SOTA systems given the complete audio waveform.

\bibliographystyle{acl_natbib}
\bibliography{paper}

\clearpage
\appendix
\section{Appendix}
\label{sec:appendix}
\subsection{Case Study}

\subsubsection{Streaming Translation }

In Table~\ref{tab:case_study}, we show an example of simultaneous decoding for cascaded systems and end-to-end systems. The cascade system has the drawbacks of error accumulation and delay accumulation. While the end-to-end model has inherent advantages in this respect. For example, in this example, our method can attend to the speaker's prosody information from the original audio input, such as pauses, so it can accurately predict the punctuation in the target language text.

\subsection{Effects of MSM}
\subsubsection{Performance on ASR}
We also validate the effects of MSM with FBANK as input on the ASR task on MuST-C EN-DE tst-COMMON set. The results are shown in Table~\ref{tab:effect_msm_asr}. There is also a performance improvement of 1.5 points, indicating that the integrate-and-fire model indeed plays an important role in learning encoded shrunk acoustic representation.

\begin{table}[!h]
    \centering
    \small
    \begin{tabular}{lc}
    \toprule
        Model & ASR (WER $\downarrow$)\\
        \midrule
       SpeechTransformer  & 15.95 \\
       \textbf{SpeechTransformer w/ MSM}  & \textbf{14.48}\\
        \bottomrule
    \end{tabular}
    \caption{Results of ASR models on MuST-C EN-DE tst-COMMON set.}
    \label{tab:effect_msm_asr}
\end{table}

\subsubsection{Training Time}
\label{sec:training_time}
Our method can be trained in parallel without the help of wait-$k$ strategy, which observably improves training efficiency. And the integration mechanism of MSM module can effectively reduce the output length of the encoder, which can reduce memory usage and increase training batch size. We experiment with FBANK feature as input on MuST-C EN-DE data set. The training time (4 Tesla-V100) for different structures is shown in Table~\ref{tab:training_time}. 

\begin{table}[!h]
    \centering
    \small
    \begin{tabular}{lc}
    \toprule
        Model & Training Time $\downarrow$\\
        \midrule
       SpeechTransformer w/ wait-$k$  & ~18 hours \\
       SpeechTransformer w/ MSM  & \textbf{~13 hours}\\
        \bottomrule
    \end{tabular}
    \caption{Training time of ST models on MuST-C EN-DE data set.}
    \label{tab:training_time}
\end{table}

\subsection{Numeric Results for Figures}

~~~
\begin{table}[!h]
    \centering
    \small
    \begin{tabular}{cccccc}
    \toprule
    \multicolumn{6}{l}{SimulST}\\
    \midrule
    \textbf{BLEU}    & 0.25 & 3.60 & 10.88&14.17 & 15.99\\
     \textbf{DAL}    & 930 & 1500& 2946& 4361& 5271\\
     \textbf{AP}    &0.37 & 0.62& 0.88& 0.96& 0.99\\
     \textbf{AL}    &604 &1097 & 2165& 2774& 3049\\
    \midrule
    \multicolumn{6}{l}{MoSST}\\
    \midrule
    \textbf{BLEU}    & 1.35 & 6.75 & 16.34& 19.77& 19.97\\
     \textbf{DAL}    & 642& 1182&2263 &3827 &4278 \\
     \textbf{AP}    & 0.29&0.53 &0.79 & 0.93& 0.96\\
     \textbf{AL}    & 208& 818& 1734& 2551& 2742\\
    \bottomrule
    \end{tabular}
    \caption{Numeric results for Figure~\ref{fig:bleu_latency}.}
    \label{tab:numeric_bleu_latency}
\end{table}

\begin{table}[!h]
    \centering
    \begin{tabular}{cccccc}
    \toprule
    \multicolumn{6}{l}{Stride=280ms}\\
    \midrule
    \textbf{BLEU}    &3.43 & 5.12& 6.87& 11.17& 13.66\\
     \textbf{DAL}    &900 & 1166&1426 & 2142& 2741\\
    \midrule
    \multicolumn{6}{l}{Stride=320ms}\\
    \midrule
    \textbf{BLEU}    & 4.93 & 7.15 & 9.08& 13.27& 15.65\\
     \textbf{DAL}    &1036 &1339 &1628 & 2415& 3041\\
    \midrule
    \multicolumn{6}{l}{Stride=400ms}\\
    \midrule
    \textbf{BLEU}    & 8.59&11.07 & 13.27& 16.65& 18.28\\
     \textbf{DAL}    &1333 &1701 &2042 & 2928& 3568\\
    \midrule
    \multicolumn{6}{l}{Stride=480ms}\\
    \midrule
    \textbf{BLEU}    &12.34 &14.60 & 16.11& 18.58& 19.49\\
     \textbf{DAL}    &1688 &2098 & 2480& 3403& 4023\\
    \bottomrule
    \end{tabular}
    \caption{Numeric results for Figure~\ref{fig:prefix_decision}.}
    \label{tab:numeric_prefix_decision}
\end{table}

\begin{table}[!ht]
    \centering
    \small
    \begin{tabular}{cccccc}
    \toprule
    \multicolumn{6}{l}{pre-fix decision}\\
    \midrule
    \textbf{BLEU}    & 4.93 & 7.15 & 9.08& 13.27& 15.65\\
     \textbf{DAL}    &1036 &1339 &1628 & 2415& 3041\\
    \midrule
    \multicolumn{6}{l}{adaptive decision}\\
    \midrule
    \textbf{BLEU}    & 7.07&9.04 &11.52 & 16.44&17.31 \\
     \textbf{DAL}    & 992& 1142&1332 &2244 & 2795\\
    \bottomrule
    \end{tabular}
    \caption{Numeric results for Figure~\ref{fig:adaptive_decision}.}
    \label{tab:numeric_adaptive_decision}
\end{table}

\begin{table*}[!h]
\small
    \centering
    \setlength{\tabcolsep}{1.5mm}{
    \begin{tabular}{lllllllllllllll}
    \toprule
     & 1 &2&3&4&5&6&7&8&9&10&11&12&13&14\\
     En (Source) & If &you & have &something& to &give&,&give&it&now&.&&&\\
     De (target) & Wenn &Sie& etwas &zu& geben& haben& \textcolor[rgb]{1,0,0}{,}&geben&Sie&es&jetzt&.&&\\
    \midrule
     ASR  & If &you&have&something&to&give&and&give&it&now&.&&&\\
     Cascades  &  &&&Wenn&Sie&etwas&zu&geben&haben&\textcolor[rgb]{1,0,0}{und}&es&jetzt&\textcolor[rgb]{1,0,0}{geben}&.\\
     \midrule
     \method &  &&&Wenn&Sie&etwas&geben&,&geben&Sie&es&jetzt&.&\\
    
    \bottomrule
    \end{tabular}}
    \caption{An example from the test set of MuST-C En-De dataset. ``ASR'' means a streaming system with 440ms' waiting latency. ``Cascades'' means a streaming pipeline contains ASR (wait-$440$ms) and NMT (wait-$3$). \method represents \method with the pre-fixed decision and wait-$3$ strategy.}
    \label{tab:case_study}
\end{table*}

\begin{table*}[!t]
    \centering
    \small
    \begin{tabular}{ccccccccc}
    \toprule
       System  & $k=1$ & $k=2$ & $k=3$ & $k=4$ & $k=5$ & $k=6$ & $k=7$&$k=\text{inf}$\\
     \midrule
       ASR (WER$\downarrow$)  & 51.9 & 43.9& 42.1&40.6 &39.6&39.0&38.7&16.25\\
    
        MT (BLEU$\uparrow$) & 17.11& 19.68& 22.80& 24.93& 26.44&27.20&27.83&31.28\\
       Cascaded ST (BLEU$\uparrow$)& 9.72& 11.24& 12.74& 13.92& 14.59&15.22 &15.51&17.60\\
    \bottomrule
    \end{tabular}
    \caption{Results of Cascaded Systems on MuST-C EN-DE tst-COMMON test set. For $k=\text{inf}$, the streaming model degrades into an offline model without beam search decoding strategy. Here the ASR model is based on SpeechTransformer~\cite{dong2018speech}, and the MT model is based on Transformer~\cite{vaswani2017attention}. Since the delays of cascaded system involving speech recognition and text translation modules are more complicated, latency metrics are not reported here. Note that cascaded ST is consisted of streaming ASR (wait-440ms, segment-40ms) and streaming MT system.}
    \label{tab:cascaded}
\end{table*}

\subsection{Compared with Cascaded System}

We build a streaming cascaded system as a baseline system by cascading a streaming speech recognition model and a text translation model. Note that the transcription generated by ASR system in the cascade streaming system is also uncorrectable. The results are shown in Table~\ref{tab:cascaded}. The error accumulation problem of the cascade system still exists in the streaming model. Compared with the results of the cascaded system, \method also has obvious performance advantages in terms of quality metrics. 

\end{document}